# Class Mean Vectors, Self-Monitoring and Self learning for Neural Classifiers

Eugene Wong
University of California at Berkeley

**Abstract**

In this paper we explore the role of sample mean in building a neural network for classification. This role is surprisingly extensive and includes: direct computation of weights without training, performance monitoring for samples without known classification, and self-training for unlabeled data. Experimental computation on a CIFAR-10 data set provides promising empirical evidence on the efficacy of a simple and widely applicable approach to some difficult problems.

Introduction

We define a one layer K-class *neural classifier* as follows: For $i = 1,2,...,K$, let $w_i \in R^n$ denote a set of weights for class $i$. For $x \in R^n$ let $w_i \cdot x$ denote *inner product* and define

$$P(i|x) = e^{w_i \cdot x} / \sum_{j \leq K} e^{w_j \cdot x} \qquad (1)$$

as the probability that an unknown vector $x$ belongs to class $i$. Classification is achieved by choosing the class with the maximum probability. The problem of designing a neural classifier is to determine a set of weights that result in highest accuracy.

In neural network parlance "learning" and "training" refer to iterative weight adjustment procedures that continually, or nearly continually, improve accuracy of classification.

Supervised Learning and Gradient Descent

A widely used iterative procedure is known as *supervised learning*. In this approach one starts with a set $T \subset R^n$ such that class membership for each of its elements is known. Let $c(x), x \in T$, denote the class label for each element $x$ in $T$. We call $T$ a *training set*. A common form of supervised learning uses *gradient descent* (GD) that begins with the choice of a loss function. For classification a very good choice is the *cross-entropy function* [Zh 2018] defined by

$$F(T) = \sum_{x \in T} f(x) \qquad f(x) = -LnP(c(x)|x) \qquad (2)$$

For simplicity and clarity, the dependency of $P(i|x), F(T)$, and $f(x)$ on the weights is suppressed in their notational expression.

Let $\nabla_i$ denote gradient with respect to $w_i$. The basic form of gradient descent provides a weight adjustment formula given by

$$\Delta w_i = -\beta \nabla_i F(T) \tag{3}$$

where β is a small positive constant and Δ denotes the change at each step of the iteration. To a first approximation we have

$$\begin{aligned}\Delta F(T) &\cong -\beta \sum_i (\nabla_i F(T)) \cdot \Delta w_i \\ &= -\beta \sum_i \|\nabla_i F(T)\|^2 \leq 0\end{aligned} \tag{4}$$

Thus, to a first approximation the loss descends at each step.

Using (2) in (3), we find

$$\begin{aligned}\Delta w_i &= -\beta \sum_{x \in T} \nabla_i f(x) \\ &= \beta \sum_{x \in T} \nabla_i [Ln P(c(x)|x)] \\ &= \beta \sum_{x \in T} [\delta_{ic(x)} - P(i|x)] x\end{aligned} \tag{5}$$

The first term in the last summation in (5) can be rewritten as

$$C_i = \sum_{x \in T} (x | c(x) = i) \tag{6}$$

We refer to $C_i$ as the *Class Mean Vector*, or simply *C-Vector*, for class $i$. We note that (6) is an un-normalized sample average for the training vectors in class $i$. We further note that $\{C_i\}$ do not depend on any of the weights and that collectively for all the classes they represent the only information on class identification that is used in the gradient descent algorithm for the cross-entropy loss function. Thus in supervised learning the supervision comes mostly from the $C_i$'s.

We now rewrite (5) in matrix form. To do so, define:

>Column $i$ of **W** is $w_i$
>Column $i$ of **P** is $\{P(i|x), x \in T\}$
>Each column of **X** is an element of $T$

We can now write (5) as

$$\Delta \boldsymbol{W} = \beta[\boldsymbol{C} - \boldsymbol{XP}] \tag{7}$$

Equation (7) is the basic formula for gradient descent expressed explicitly in terms of the constituents of the gradient.

<u>Direct Computation of Weights</u>

In addition to its role in gradient descent, **C** can also be used in direct computation of weights. To begin with, setting

$$w_i = C_i / \|C_i\| \tag{8}$$

is a reasonable first choice for the weights. Using this choice in the neural classifier classifies any sample vector $x$ by minimizing its distance to $C_i/\|C_i\|$. As such it is a linear classifier and a very old method for designing neural classifiers. Effectively, the resulting neural network is a linear SVM. [Su 1999]

A second method for direct computation of the weights is by linearizing (7). We do this by replacing $P(i|x)$ in (5) by the inner-product $w_i \cdot x$. The logic for doing this is that the maxima of the two quantities $P(i|x)$ and $w_i \cdot x$ occur at the same $i$. Now (7) takes the form

$$\Delta \mathbf{W} = \beta[\mathbf{C} - \mathbf{XX'W}] \tag{9}$$

where prime denote transpose.

Since **C** and **XX'** stay fixed as the weight change progresses, (9) is a linear dynamical equation in **W**. As such, (9) has an explicit solution in either discrete-time or continuous-time form. For example, in continuous-time form the solution is given by

$$\mathbf{W}(t) = \exp(-\beta \mathbf{XX'}t)[\mathbf{W}(0) - (\mathbf{XX'})^{-1}\mathbf{C}] + (\mathbf{XX'})^{-1}\mathbf{C} \tag{10}$$

In either case the solution converges to

$$\mathbf{W} = (\mathbf{XX'})^{-1}\mathbf{C} \tag{11}$$

Equations (8) and (11) are two explicit ways in which the weights can be computed, rather than learned. The computation time for $(\mathbf{XX'})^{-1}$ is highly dependent on the dimension of the sample vectors. For modest dimensions, say $\leq 1000$, the time is much less than that of a reasonable execution of gradient descent.

As an indication of the performance of these two methods of direct computation, an experiment using 3000 labeled samples of the CIFAR-10 data set was undertaken. The results are summarized below:

| Method | Accuracy | |
|---|---|---|
| **C** | .464 | |
| $(\mathbf{XX'})^{-1}$ | .7 | |
| Grad. Des. | .661 | (400 iterations with .3% learning rate) |

It is interesting to note that using the mean vectors for weights produces relatively modest results, but direct computation by linearizing the gradient results in an accuracy better than that achieved by gradient descent in a sequence of 400 iterations..

We also note in passing that neither method for direct computation of weights leads to a low value for the loss function. Indeed, in either case the loss is scarcely less than that for random weights. It is fair to say that low loss means good accuracy, but the converse is not true. In this sense the cross entropy loss function is far from perfect. This is a surprising outcome.

E Marker: A Means for Monitoring Performance

During the operational mode of a trained neural classifier, there is no existing method to monitor its accuracy, because class label is not available. No matter how good the weights may be initially, performance can deteriorate and one would be unaware of any such performance degradation. Thus, there is a need for finding a monitoring tool that can measure performance without knowing the class label. Class Mean Vectors provides a way.

Let $S$ denote a set of unlabeled vectors from the same population as the training set $T$. Let **C** denote the C Vectors generated by $T$. Now define

$$M_i = \sum_{x \in S} \{x: Max_j\,[P(j|x)] = P(i|x)\} \tag{12}$$

In other words $M_i$ is the sum of the vectors in $S$ for which the maximum probability occurs at class $i$. Now define

$$E_i^2 = \left(\tfrac{1}{4}\right)\left\|\frac{C_i}{\|C_i\|} - \frac{M_i}{\|M_i\|}\right\|^2 \tag{13}$$

as a measure of error for class $i$. Both terms on the right hand side of (13) have been normalized to be unit vectors. This normalization automatically adjusts for any scale, sample size, and class distribution differences between $S$ and $T$. Since both terms in (13) are unit vectors, we can define the angle $\vartheta_i$ between them by

$$\text{Cos}(\vartheta_i) = (C_i/\|C_i\|) \cdot (M_i/\|M_i\|) \tag{14}$$

Equation (13) can now be evaluated and written in full as

$$E_i^2 = \sin^2\left(\frac{\vartheta_i}{2}\right) \tag{15}$$

We now define the *E-Marker* by

$$E = \sqrt{Avg_i(E_i^2)} \tag{16}$$

To validate the role of e-marker, we conducted the following experiment: Let $T$ comprise 3000 sample vectors from the CIFAR-10 data set, each with dimension 400 and class-labeled 0 through 9. Let $S$ comprise another 1500 such vectors, but with class-labels temporarily masked. Set the initial weights to random numbers, use the cross-entropy loss function, and apply gradient descent on the vectors in $T$. As the algorithm progresses, we observe and compare the loss function $F(T,t)$ and the e-marker $E(S,t)$, where $t$ denotes the steps in the progression of the algorithm. The results for a 400-iteration run are shown below in Figure 1.

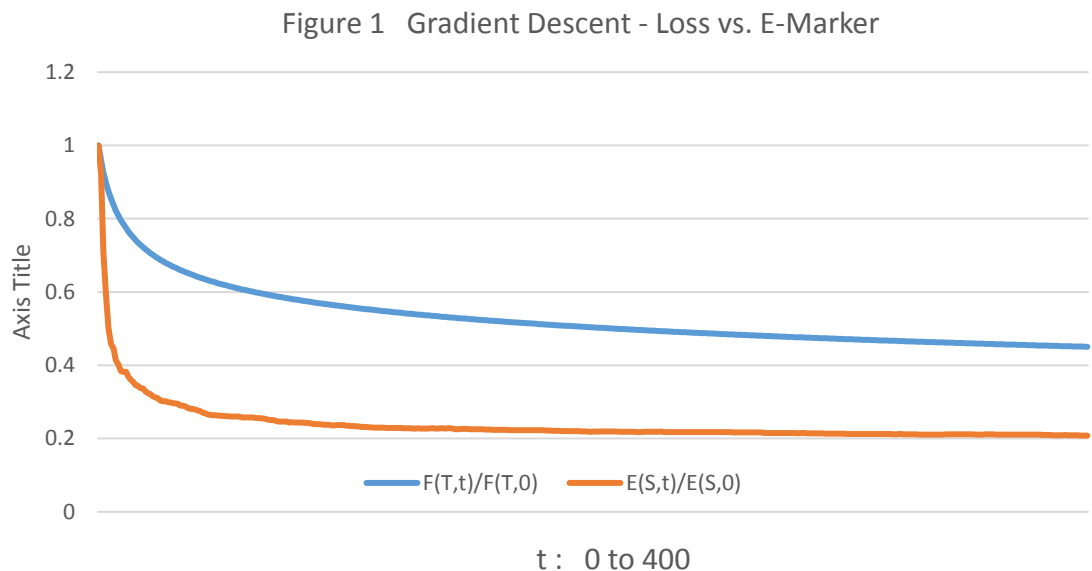

Figure 1  Gradient Descent - Loss vs. E-Marker

We note that the E-Marker not only descends, but does so much more steeply than the loss function, indicating that the weight changes are having a strong effect on the test set $S$.

The steepness of the e-marker descent is surprising, but it actually augurs well the performance on $S$. To see that, we now unmask the labels on set $S$ and calculate the accuracy $P(S,t)$. To get the best visual comparison, we plot $Ln(\frac{P(S,t)}{P(S,0)})$ and $-Ln(\frac{E(S,t)}{E(S,0)})$ against $t$ for a 400 iteration run of gradient descent. The result is presented in Figure 2.

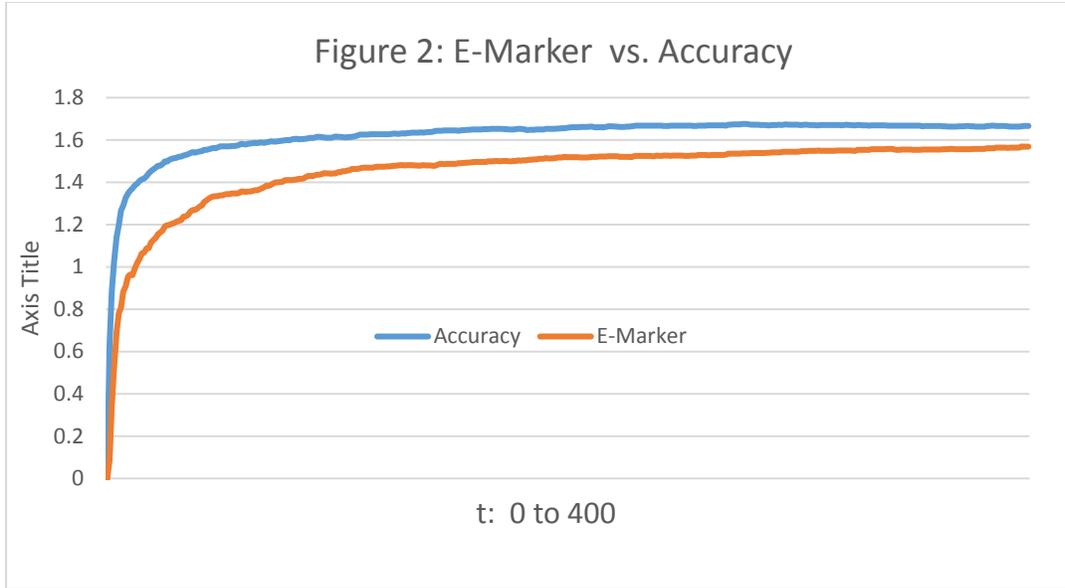

Figure 2: E-Marker vs. Accuracy

The tracking shown in Figure 2 is remarkable for a performance marker that uses no labeling information at all.

Pseudo Gradient Descent and Self-Learning

The performance of a neural classifier can be monitored continuously using a moving-window form of the E-Marker. But what can one do if the E-Marker shows a degradation in performance? Once again the C - Vectors **C** provide an answer. The answer is to use **C** to construct a pseudo-gradient to enable self-learning by data with unknown classification.

Let $S$ be a set of sample vectors with unknown classification. Let **C** be a set of C Vectors derived from one or more training sets distinct from $S$. Let the elements of $S$ form the columns of matrix **Z**, and let **P** be a matrix with columns indexed by the classes and with rows indexed by the sample vectors of $S$. We define a pseudo gradient $\nabla$ by

$$\nabla_i = C_i/\|C_i\| - (\mathbf{ZP})_i \qquad (17)$$

and a weight change formula

$$\Delta w_i = \beta[C_i/\|C_i\| - (\mathbf{ZP})_i] \qquad (18)$$

For the initial values of the weights we have several choices:

(a) minimum distance $w_i = C_i/\|C_i\|$
(b) linearization $w_i = (\mathbf{ZZ'})^{-1}(C_i/\|C_i\|)$
© random

The pseudo gradient descent algorithm was applied to a data set $S$ consisting of 1500 sample vectors from CIFAR-10 with class identification temporarily masked, using Class Mean Vectors derived from 3000 similar but distinct CIFAR-10 samples. All three choices for initial weight values were tried. The results are shown in Figure 3.

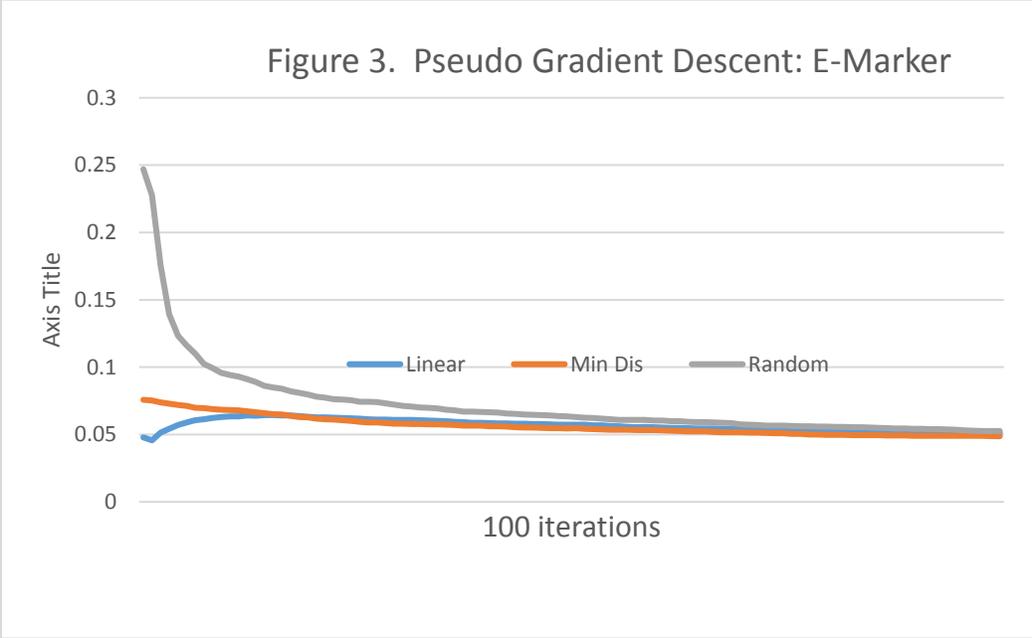

We note that although the E-Marker values for different initial weights start very differently, they quickly converge to near identical, and low, numbers, indicating that pseudo gradient descent has worked in this case. That is verified by unmasking the class labels on $S$ and observing the actual accuracy values. The results are shown in Figure 4.

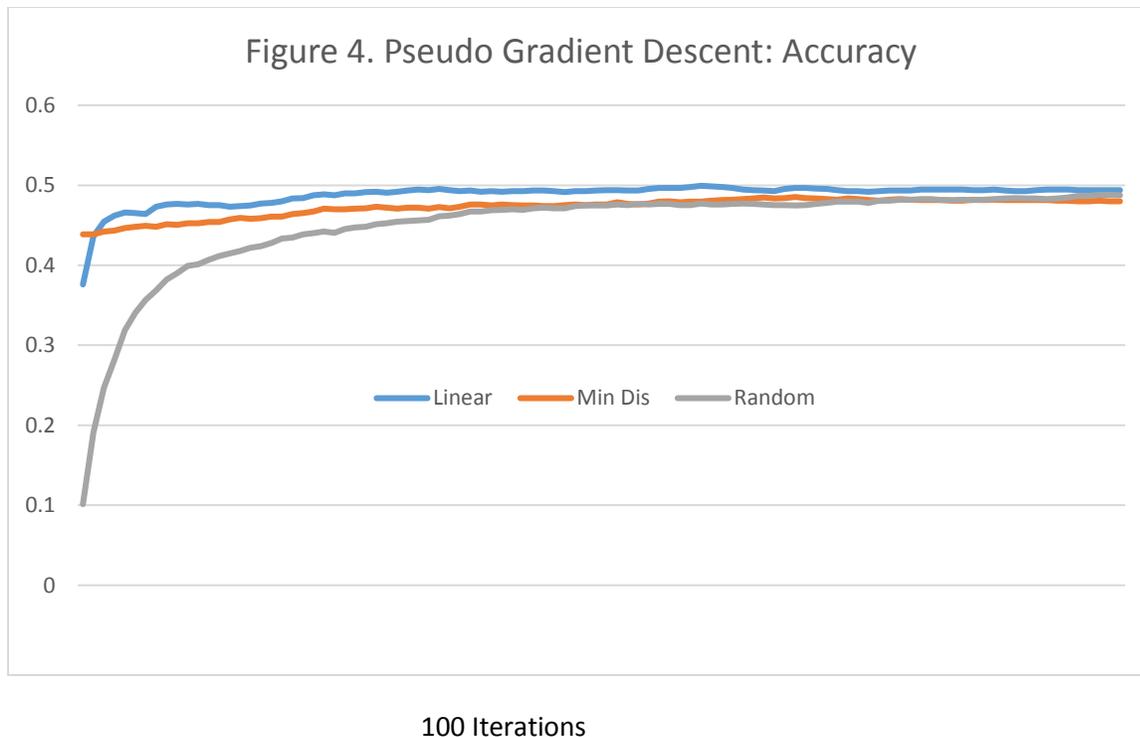

100 Iterations

Again, we see a convergence to the same value irrespective of the choice of initial weights. The final value of accuracy is quite good and validates the concept of pseudo gradient descent.

C Vectors and Zero Shot Learning

Let $T$ be a set of labeled data vectors from classes in $Cl(T)$ and let $S$ be a set of unlabeled data vectors of the same dimension from classes in $Cl(S)$. We assume that $Cl(T)$ and $Cl(S)$ are disjoint. The problem of Zero Shot Learning is to classify the vectors of $S$ by learning from $T$. Simple logic suggests that without additional side information, this is an impossible task. (See e.g., [Ak 2017] for a large variety of possible side information.)

A popular existing approach to the problem might be called "*attribute compatibility learning*."(See e.g., [Xi 2019])  We provide a simplified version as follows: The side information in this approach is obtained by characterizing the combined collection of classes $Cl = Cl(T) \cup Cl(S)$ by a set of functions $\phi$ called *attributes*. Compatibility learning is then done in two stages. The first stage consists of a neural network that has the elements $x$ of $T$ as input and its output $y(x)$ is matched to $\phi(c(x))$ where $c(x)$ denotes the class label of $x$. The matching is done using some learning algorithm (e.g., GD) via some loss function. A number of choices for the loss function have been suggested. The second stage is a neural classifier that classifies the output of the first stage with or without additional learning. For example, it can be simply a linear SVM that classifies according to the maximum inner product between the output of the first stage and the attribute vectors of the classes. In practice an important additional step of mapping the input vectors of $T$ and $S$ into "features" is often necessary to achieve a good compatibility between the input and the attributes.

The C Vector approach to the problem is rather different. As in self-learning, we extract the C Vectors from $T$ and construct a pseudo gradient on $S$. We can now use pseudo gradient descent to adjust the weights and monitor progress with the E Marker. The side information we need is some kind of correspondence between $Cl(T)$ and $Cl(S)$. We have tried two specific choices:

(1) correlation function $R(a, b)$, $(a, b) \in Cl(T) \times Cl(S)$
(2) class mapping $\rho: Cl(T) \to Cl(S)$

The mapping $\rho$ should be "onto," lest some of the classes of $Cl(S)$ are left out of the classification.

At the end of pseudo gradient descent the output of the neural classifier is of the form: $p(x, a), (x, a) \in S \times Cl(T)$. With assumption (1) we define

$$\pi(x, b) = \sum_a R(a, b) p(x, a) \tag{19}$$

Classification is achieved by maximizing $\pi$. With assumption (2) we first maximize $p(x, a)$ and define $a_m$ by

$$p(x, a_m) = max_a p(x, a) \tag{20}$$

Then $x$ is assigned to the class $\rho(a_m) \in Cl(S)$.

As an experiment, 4500 samples from CIFAR10 were used. These samples are labeled 0 to 9. We set

$$Cl(S) = \{0 \text{ to } 4\} \text{ and } Cl(T) = \{5 \text{ to } 9\}$$

A correlation function $R(a, b)$ and a one-to-one map $\rho$ were found using the C vectors from both $S$ and $T$.

The C – Vectors were computed from $T$ and then the vectors of $S$ were classified in two ways:
   (a) The normalized C Vectors were used as weights in a neural classifier, followed by an assignment of classes using $R$ and $\rho$ per earlier discussion.
   (a) The normalized C Vectors were used in 200 iteration pseudo gradient descent run, again followed by an assignment of classes using $R$ and $\rho$.

The classification accuracy for the vectors of $S$ for the different methods was found as follows:

| Weights | R | $\rho$ |
| --- | --- | --- |
| Norm. C Vectors | .49 | .45 |
| Pseudo GD | .45 | .42 |

Surprisingly, the use of C Vectors as weights outperformed the pseudo GD.

As a first attempt the results of the C Vector approach to Zero-Shot Learning appear promising. Aside from performance, this approach offers a number of distinct advantages. These include: learning not only from training data, but also test data; a minimal need for side information; the ability to gauge effectiveness through E Marker monitoring; and its flexibility to combine and complement existing approaches such as attribute based Zero Shot learning.

Summary

Given a set $T$ of class-labeled samples, we define C Vector for each class as the sum of the vectors in $T$ that belong to that class. Clearly, C Vectors are nothing more than un-normalized sample averages. Surprisingly, they play a rather extensive role both inside and outside $T$. Using C Vectors in performance monitoring and pseudo gradient descent on unlabeled sets is a particularly novel and possibly important application.

**Acknowledgement**

I am grateful to Dr. J. M. Ho of Academia Sinica and Professor C. Y. Lee of the National Chiao Tung University (NCTU) in Taiwan for many helpful comments. Eugene Lee of the NCTU team provided me with the CIFAR-10 data, suggested the Zero Shot application and was helpful to me throughout the work reported in this paper.